\documentclass[sigconf, nonacm]{acmart}

\AtBeginDocument{%
  }

\usepackage{graphicx}
\usepackage{xcolor}
\usepackage{algorithm}
\usepackage{algorithmic}
\usepackage{tabularx}
\usepackage{color}
\usepackage{xspace}
\usepackage{subcaption}

\newcommand{\util}{\mathrm{util}}
\def\nngpso{NNGPSO\xspace}
\def\cnngpso{CNNPSO\xspace}
\def\dnngpso{DNNPSO\xspace}
\def\Particles{\mathcal{P}}

\usepackage{bm}
\def\Pos{{\bm x}}

\def\GlobalBest{{\bm g}}
\def\ActualGB{{\bm a}}
\def\Velocity{{\bm v}}

\begin{document}

\title[Deep Neural Network-guided PSO for Tracking a Global Optimal Position in Complex Dynamic Environment]{Deep Neural Network-guided PSO for Tracking a Global Optimal Position in Complex Dynamic Environment}

\author{Stephen Raharja}
\email{stephen.r@fuji.waseda.jp}
\orcid{0009-0002-8163-4357}
\affiliation{%
  \institution{Waseda University}
  \city{Shinjuku City}
  \state{Tokyo}
  \country{Japan}
}

\author{Toshiharu Sugawara}
\email{sugawara@waseda.jp}
\orcid{0000-0002-9271-4507}
\affiliation{%
  \institution{Waseda University}
  \city{Shinjuku City}
  \state{Tokyo}
  \country{Japan}
}


\begin{abstract}
We propose novel {\em particle swarm optimization} (PSO) variants incorporated with deep neural networks (DNNs) for particles to pursue globally optimal positions in dynamic environments. PSO is a heuristic approach for solving complex optimization problems. However, canonical PSO and its variants struggle to adapt efficiently to dynamic environments, in which the global optimum moves over time, and to track them accurately. Many PSO algorithms improve convergence by increasing the swarm size beyond potential optima, which are global/local optima but are not identified until they are discovered. Additionally, in dynamic environments, several methods use multiple sub-population and re-diversification mechanisms to address outdated memory and local optima entrapment. To track the global optimum in dynamic environments with smaller swarm sizes, the DNNs in our methods determine particle movement by learning environmental characteristics and adapting dynamics to pursue moving optimal positions. This enables particles to adapt to environmental changes and predict the moving optima. We propose two variants: a swarm with a centralized network and distributed networks for all particles. Our experimental results show that both variants can track moving potential optima with lower cumulative tracking error than those of several recent PSO-based algorithms, with fewer particles than potential optima. 
\end{abstract}


\maketitle


\section{Introduction}
 Climate-related disasters have increased over the past two decades due to intensifying global warming~\cite{UNDRR2020_HumanCostDisasters}. The growing concentration of the population in urban areas has heightened the complexity of disaster response. As urbanization has increased the population density in cities and suburban sprawl, potential disaster impacts have multiplied~\cite{Feng2021UrbanizationIO, Guo2024GlobalEO}. With urbanization projected to grow in the coming decades~\cite{UN_DESA_2019}, more efficient and adaptive search-and-rescue (SAR) strategies are needed. 
\par

Recent technological progress has enabled the development of self-driving intelligent tools, such as {\em autonomous unmanned   aerial vehicles} (UAVs) and drones, for multiple uses across industries~\cite{Otto2018OptimizationAF}. The cost of drone hardware continues to decline, while the power consumption and production cost of modern processors have decreased~\cite{Nordhaus2007TwoCO}, enabling better computational capabilities in drones. The convergence of affordability, mobility, and computational power renders drone-based distributed sensing systems increasingly practical for large-scale SAR frameworks. In post-disaster scenarios, survivors often migrate collectively towards important locations, such as community centers, hospitals, or pre-designated evacuation sites~\cite{Bengtsson2011ImprovedRT}. However, even with drones strategically pre-positioned in high-density urban regions, anticipating where survivors will gather and how these clusters evolve remains a challenge.
\par

The migration patterns of survivors can be simulated to predict disasters~\cite{Song2016PredictionAS}; therefore, they can be modeled as a {\em dynamic optimization problem} (DOP), where survivor clusters move in the environment over time. Because the main objective of SAR agencies is to save the most lives by discovering areas with the most survivors~\cite{fema_disaster_plan}, this study aimed to identify and track areas with the highest survivor concentration in a limited geographical region within a city, town, or village. Other optimization problems, such as job scheduling~\cite{pso_job_scheduling} and vehicle routing~\cite{pso_vehicle_routing}, can also be modeled as DOPs with the objective of identifying and tracking a single global optimum.
\par

Several studies have proposed PSO-based algorithms for solving DOPs~\cite{SPSO,PSPSO,DMPSO}. However, these PSO algorithms assume that the environment changes at intervals rather than continuously, such as in the {\em generalized moving peaks   benchmark}~\cite{GMPB}. Furthermore, PSO algorithms require a particle population larger than the number of potential optima, which cannot be identified as local or global until they are discovered by the swarm.
\par

Therefore, we propose a PSO-based method for DOPs, where drones are modeled as particles that identify and track the global optimal position over time. Unlike conventional PSO algorithms, which assume periodic changes with abundant swarm members, our method is designed for dynamic environments with fewer particles than potential optima in continuously changing environments.
\par

First, the problem is formalized by reviewing SAR scenarios.  Next, the proposed method, {\em neural network-guided PSO} (\nngpso), is described.  NNGPSO leverages {\em deep neural networks} (DNNs) to guide particle movement, enhancing tracking of regions with the highest estimated survivor density over time. The method was experimentally evaluated and compared across multiple dynamic scenarios. Compared with canonical particle swarm optimization (PSO) and recent PSO-based techniques\cite{SPSO,PSPSO}, NNGPSO tracks moving global optima more effectively, even with fewer particles in dynamic optimization problems (DOPs). These results suggest that NNGPSO can improve drone-assisted SAR operations when the number of drones is smaller than the number of potential survivor clusters, as the DNN can efficiently guide particles (drones) toward promising areas. 
\par

\section{Related Work}
Canonical PSO~\cite{canon_pso} is a population-based stochastic optimization method inspired by swarming creatures, such as bee colonies and bird flocks. A swarm of particles explores the search space by updating the positions based on their experience and shared information. PSO algorithms and their variants are popular because of their simplicity, low computational cost, and flexibility in tackling nonlinear, multi-modal, multi-issue, and high-dimensional optimization problems. This foundation has led to the development of many PSO variants incorporating adaptive parameters, hybridization with other metaheuristics, and multiple interacting swarms \cite{Poli2009EditorialFT}. However, most conventional PSO algorithms are designed for static environments and are unsuitable for DOPs in gradually changing environments.
\par

Meanwhile, difficulties in solving DOPs limit the usability of conventional PSO-based algorithms. One issue is outdated memory, where previously optimal solutions guide the swarm after the fitness landscape shifts, causing the search to focus on obsolete areas. The convergence of the swarm around prior optima leads to diversity loss, reducing the exploratory capacity. Consequently, the swarm cannot discover new promising areas, hampering its adaptability to dynamic environments~\cite{GMPB}.
\par

Various techniques have been developed to address the difficulties in dynamic environments. For outdated memory, one technique forces the swarm to explore regions different from the current converged positions, allowing it to refresh the outdated memory and explore new regions. Particle re-diversification can be achieved by adjusting positions using random perturbation~\cite{DMPSO}, re-initializing particles in new regions~\cite{SPSO}, or replacing converged particles with new ones~\cite{PSPSO}.
\par

The loss of diversity in swarms is a challenge in dynamic environments. Some algorithms~\cite{SPSO, PSPSO, DMPSO} use re-diversification to re-initialize particles in new positions to allow the rediscovery of explored or new regions. Another technique involves distancing particles and sub-swarms to promote exploration by spreading them wider. Particles can be distanced using a repulsion force~\cite{charged_pso}, but this may prevent convergence owing to strong repulsion. Some algorithms~\cite{SPSO, PSPSO} employ a convergence threshold radius to replace worse sub-swarms with new ones in different regions.
\par

Some algorithms employ explicit environmental change detection to improve adaptation~\cite{SPSO}. Explicit change detection enables granular control of triggering mechanisms, such as particle re-diversification, potentially allowing quicker shifts from exploiting optimal positions to exploring when the environment changes. However, our experiment showed that detecting changes often negatively impacts convergence when the environment changes too frequently, preventing the swarm from exploiting the optimal positions.
\par

Nevertheless, most experiments evaluating PSO-based algorithms in static or dynamic environments use more particles than the potential optima to allow reliable swarm convergence, with convergence improving with a larger swarm size~\cite{pso_population}. However, a smaller swarm can perform as well as a larger one without changing other parameters~\cite{minimal_pso_population}. Because the number of potential optima is likely unknown beforehand, we used the worst-case assumption that the swarm size is smaller than the potential optima count. Our method is designed for dynamic environments without assuming the frequency of environmental changes or number of optimal points. Specifically, it is a method for reliably exploring a changing dynamic environment with more potential optimal points than particles, without explicit change detection.
\par

\begin{figure}
  \centering
  \begin{subfigure}[t]{0.7\linewidth}
    \includegraphics[width=\linewidth]{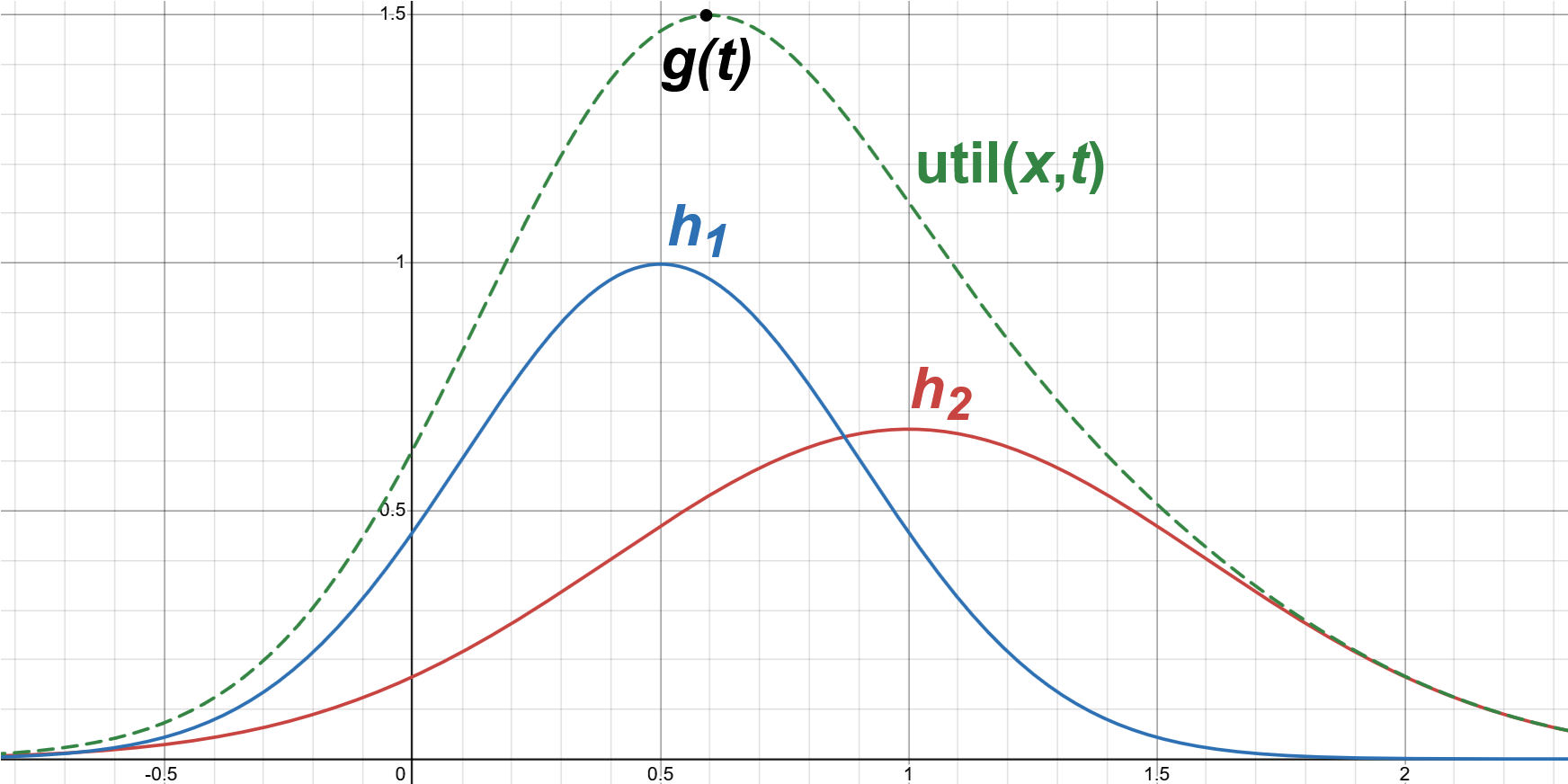}
    \caption{$g(t)$ close to the $h_1$}
    \label{fig:global_close_to_peak}
  \end{subfigure}
  \hfill
  \begin{subfigure}[t]{0.7\linewidth}
    \includegraphics[width=\linewidth]{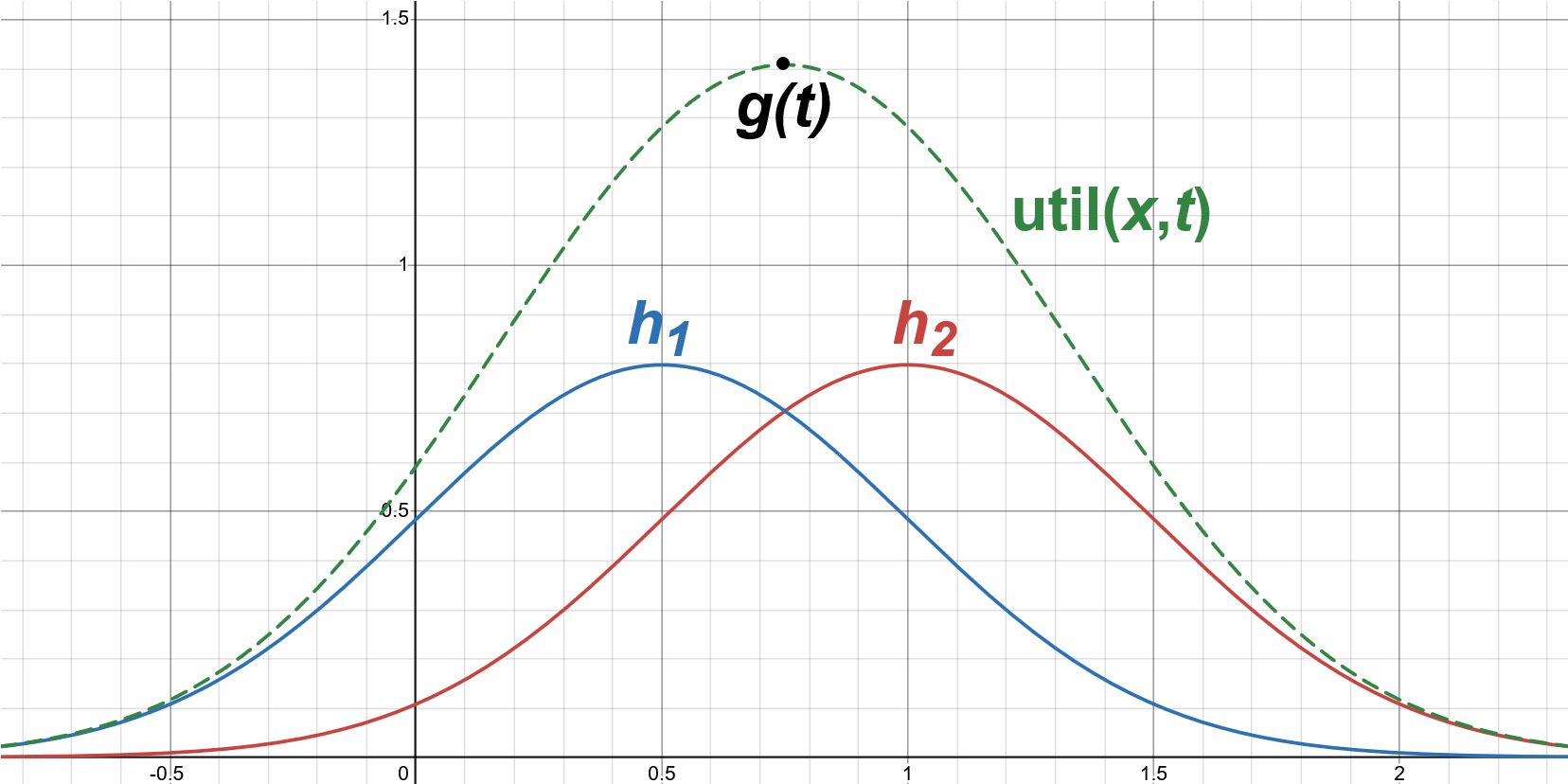}
    \caption{$g(t)$ between $h_1$ and $h_2$}
    \label{fig:global_between_peaks}
  \end{subfigure}
  \caption{Example of $g(t)$ location relative to peaks $h_1, h_2$.}
  \label{fig:peaks_global}
    \Description[]{The figure shows the example positions of a $g(t)$ relative to two peaks on 2-dimensional plane, with y
    equals to the probability density function. In the first figure,
    the first peak of $h_1$ has smaller standard deviation ,
    resulting in the position of $g(t)$ close to the center of
    $h1$. In the second figure, both peaks $h_1, h_2$ have the same
    standard deviation, and result in $g(t)$ to be positioned exactly
    between the two peaks.} 
\end{figure}

\section{Preliminaries}
\subsection{Environment Model and Problem Description}
Inspired by the characteristics of estimated search areas of human presence in post-disaster situations, we model the environment $E$, a finite square-shaped two-dimensional bounded vector space in $\mathbb{R}\times\mathbb{R}$ centered at $(0,0)$, where each dimension is bounded by $[-E_{bound},E_{bound}]$. Areas of potential survivors are abstracted into areas of potential solutions, represented as a set of peaks $H = \{h_1, \dots, h_l\}$ ($\subset E$). Moreover, socially and infrastructurally significant locations where survivors move are abstracted into final destinations for solutions, represented as a set of centers $C = \{c_1, \dots, c_m\}$ ($\subset E$).
\par

To mimic the unpredictability in post-disaster environments, the initial positions of peaks $h \in H$ and centers $c \in C$ are random points in $E$, and each center $c \in C$ is randomly assigned to a peak $h\in H$. We assume that centers $C$ remain static, while $h\in H$ may move linearly towards its assigned center $c$ with speed vector $\boldsymbol{\epsilon}$ at each discrete time $t$, representing converging areas of potential solutions into final destinations. Destinations $c$ prevent peaks from diverging beyond the environment bounds, as a random walk in finite time leads to an unpredictable result~\cite{central_limit_theorem}.
\par

A peak $h_i\in H$ denotes the location and influence area of a solution represented as a bivariate Gaussian distribution $(\bm{\mu},\bm{\sigma})$, where the mean $\bm{\mu}_i=(\mu_{1,i}, \mu_{2,i})$ denotes the location, and $\bm{\sigma}_i=(\sigma_{1,i}, \sigma_{2,i},\rho_i)$ is the tuple of standard deviations (SDs) $\sigma_1$ and $\sigma_2$ in two axioms representing the peak's influence area, with correlation $\rho_i$. For each peak $h_i$, the peak movement at time $t$ is defined as $\bm{\mu}_i(t+1) = \bm{\mu}_i(t) + \bm{\epsilon}_i(t)$, and $\bm{\sigma}_i$ remains constant.
\par

The corresponding utility function $\util(\Pos, t)$ quantifies the cumulative probability density function of all peaks $H$ at the coordinate $\Pos = (x_1, x_2)$ at time $t$, representing the strength of the solution at a specific coordinate in the environment $E$ at state $t$.  Formally, it is defined as
\begin{equation}\nonumber
  \util(\Pos, t) =
  \sum_{i=1}^{|H|}
  \frac{1}{2\pi\sigma_{1,i}\sigma_{2,i}}
  \exp\!\left(
  -\frac{1}{2}\left[
  \left(\frac{x_1 - \mu_{1,i}}{\sigma_{1,i}}\right)^2 +
  \left(\frac{x_2 - \mu_{2,i}}{\sigma_{2,i}}\right)^2
  \right]
  \right),
\end{equation}
where $\rho_i=0$ is assumed, indicating independence between the two spatial dimensions.
\par

We define the global optimal position at time $t$ as 
\begin{equation}\nonumber
\GlobalBest(t) = \mathrm{argmax}_{\Pos\in E} \util(\Pos, t)
\end{equation}
because the utility value at $x$ may differ between discrete times in a changing environment. Notably, $\util(x,t)$ changes as peaks $H$ moves at every time $t$, and it is possible that $\util(x,t)$ follows an increasing or decreasing trend. In other words, $\util(x,t-1) \lessgtr \util(x,t)$.
\par

Our problem aims to track $\GlobalBest(t)$ at each time $t$. As our assumed environment is sufficiently small that each peak $h$ can be within another peak's area of influence, the global optimal $\GlobalBest(t)$ might be near one peak (Fig.~\ref{fig:global_close_to_peak}) or between two peaks (Fig.~\ref{fig:global_between_peaks}). In the SAR context, the former indicates a smaller survivor group merging into a larger one, whereas the latter represents two equal-sized groups merging. Thus, our experiment aimed to minimize the cumulative tracking error value between the actual and detected global optimal at $t$.
\par

\subsection{Canonical Particle Swarm Optimization}\label{sect:cpso}
Let $\Particles=\{p_1, \dots, p_n\}$ be a set of $n$ particles (where $n$ is a positive integer). The velocity of particle $i\in\Particles$ in vector space is updated by combining three components: {\em inertia} (previous velocity), cognitive learning (attraction towards its personal local best position), and social learning (attraction towards the global best position). Formally, the velocity $\Velocity_i(t)$ and position $\Pos_i(t)$ of $i$ at time $t$ are updated formulas as follows:
\begin{align}
  \Pos_i(t+1)=&\Pos_i(t)  + \Velocity_i(t+1) + w \cdot \Velocity_i(t) \textrm{ and}\label{eqn:position_update}\\
  \Velocity_i(t+1)=& c_1 r_1(y_i(t)-\Pos_i(t)) 
     +c_2 r_2(\GlobalBest(t)-\Pos_i(t)),\label{eqn:c_pso}
\end{align}
where $w$ is the velocity weight, $c_1>0$ and $c_2>0$ are acceleration coefficients, $r_1$ and $r_2$ are random numbers between $0$ and $1$, $y_i(t)$ is the personal best position of particle $i$, and $\GlobalBest(t)$ is the global best position found by the swarm. Using these formulas, each particle balances exploration and exploitation. The inertia term maintains momentum, the cognitive term enables local search around successful positions, and the social term drives convergence towards the best-known global solutions~\cite{Clerc-Kennedy-2002}.
\par

\subsection{Deep Neural Networks}
A DNN is a machine learning method with multiple network layers for nonlinear transformations. Each layer extracts higher-level features from raw input data, allowing DNNs to capture complex hierarchical relationships that are difficult to represent using traditional methods~\cite{Goodfellow-et-al-2016}. DNNs are powerful because of their universal approximation properties. Under suitable conditions, they can approximate various nonlinear functions with arbitrary precision~\cite{Cybenko-1989, Lu2017TheEP}. This expressive power has led to state-of-the-art performance across domains, including computer vision and reinforcement learning~\cite{Krizhevsky2012ImageNetCW,Mnih2015HumanlevelCT}.
\par

In our context, DNNs offer a flexible framework for capturing complex patterns in dynamic environments, thereby enhancing the convergence performance of the proposed methods~\cite{Parisi2018ContinualLL}. This is valuable in our modeled problem, where the particle count is smaller than the number of potential optima, unlike typical PSO configurations, where the swarm size exceeds the optima by a large margin.
\par

\begin{figure}
  \includegraphics[width=0.70\linewidth]{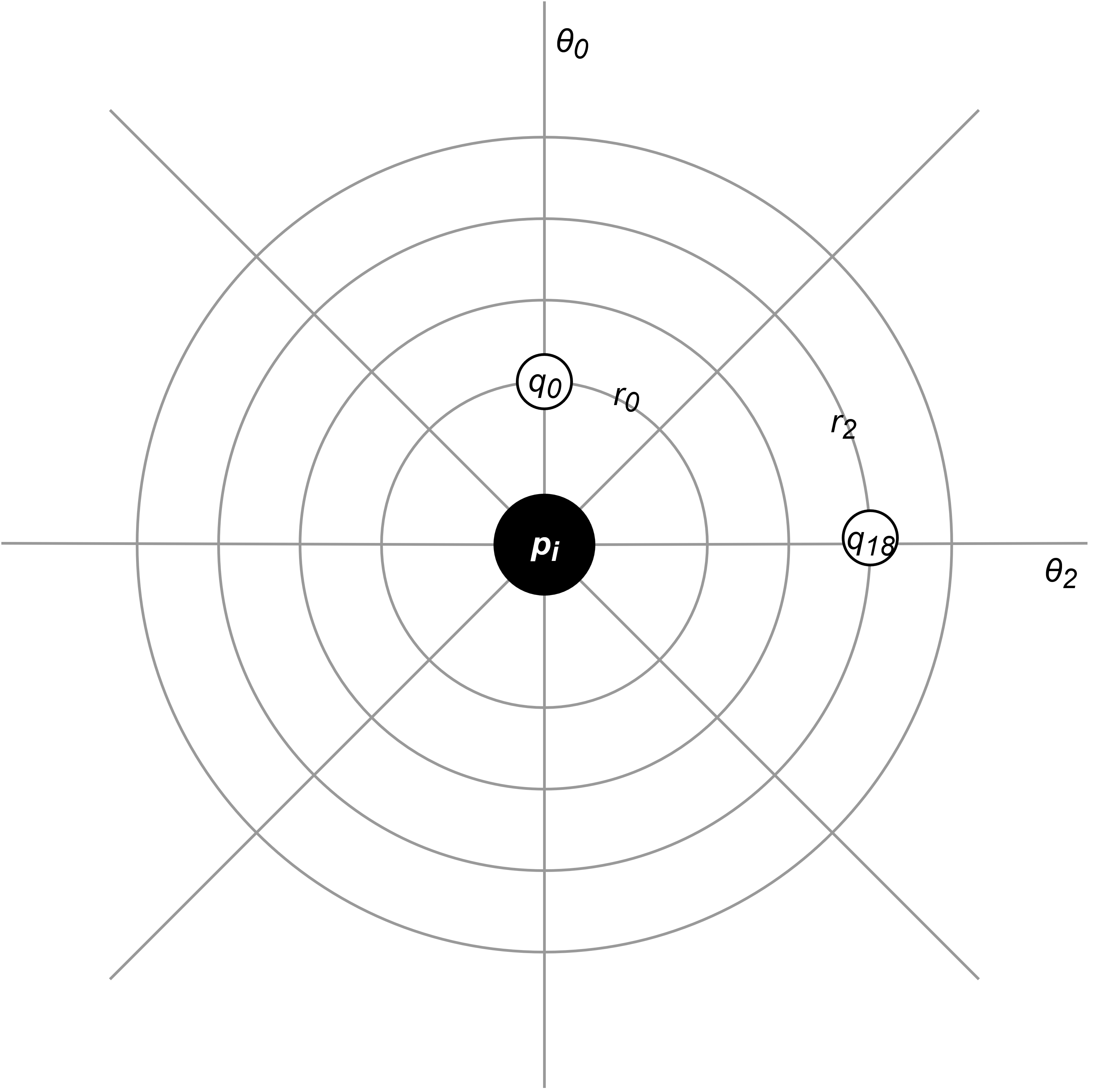}
  \caption{Visualization of particle $p_i$ in \nngpso with two example
    observation points}
      \Description[]{
Top view of a particle in \nngpso, where each particle has multiple
surrounding observation points. The position and utility values of
each observation point are used as input data for the deep neural
network.}
  \label{fig:dnnpso_particle}
\end{figure}

\section{Proposed Methods}
\subsection{Implementing DNN for PSO}
\subsubsection{Constructing Network Input $\mathcal{I}(t)$} 
To track the global optimum in a dynamic environment where the particle count is smaller than the number of potential  optima, we employ a DNN to estimate the next position $x(t+1)$ of each particle. The input data are a set of normalized coordinates and utility values from a particle in \nngpso, defined as $\mathcal{I}(t)$.
\par

To construct the normalized input set $\mathcal{I}(t)$, each particle $p_i$ is associated with an observation point $Q_i = \{q_{i,1}, q_{i,2}, \ldots, q_{i,n}\}$. Each point $q_{i,j} \in Q_i$ is defined as the intersection between an observation ring of radius $\hat{r}_{i,j} = m_i \cdot r_{i,j}$ and the direction $\theta_{i,j}$. Formally,
\begin{equation}\label{eqn:dnnpso_orbit_update}
Q_i = \{\, \hat{r}_{i,j} \cap \theta_{i,j} \mid r_{i,j} \in \bm{r}_i,\, \theta_{i,j} \in \bm{\theta}_i \,\},
\end{equation}
where $m_i$ is a scalar multiplier applied to the base radius $r_{i,j} \in \bm{r}_i$, and direction $\theta_{i,j}$ is chosen from predefined directions $\bm{\theta}_i$.
\par

Figure~\ref{fig:dnnpso_particle} shows the top view of a particle in \nngpso including two example observation points, $q_{i,0} = \theta_{i,0} \cap r_{i,0}$ and $q_{i,18}=\theta_{i,2}\cap r_{i,2}$. Each particle $\forall p_i\in\Particles$ has multiple surrounding observation points. The position of each observation point and the utility value $util(q_{i,*},t)$ are fed to the DNN as input data.
\par

The current position $\Pos_i$ and observation points $Q_i$ are flattened into one-dimensional vectors using the mapping function $\Phi(\cdot)$. For each coordinate, the spatial position components are concatenated with the utility value $\Phi(x) = [x_1',\, x_2',\,   \util(x,t)]$, where the coordinates are normalized to $[-1,1]$ by dividing by the environment boundary size $x_n' = \frac{x_n}{E_{{\it bound}}}$ for $n=1,2$, where $\Pos=(x_1,x_2)$. The complete input set for the network at time $t$ created by particle $i$ is then defined as
\begin{equation}\label{algo:dnnpso_inputset}
\mathcal{I}_i(t) =
\big[
\Phi(\Pos_i) \cup
\bigcup_j\Phi(q_{i,j})
\big],
\end{equation}
which combines the particle position and local observations into a flattened, normalized representation for DNN input.
\par

We do not incorporate the local or global best positions into the network input $\mathcal{I}(t)$. In dynamic environments, such information can be misleading, as these positions may be outdated and hinder convergence performance~\cite{Blackwell2006MultiswarmsEA}. In our experiments, omitting both the local and global best positions from the network input improved the tracking accuracy and convergence stability.
\par

\subsubsection{Calculating the Next Position}
Using the network input data $\mathcal{I}(t)$, the next position $\Pos(t+1)$ for each particle can be calculated using feedforward neural networks. Specifically,
\begin{align}
  \hat{\Pos}(t+1) &= \mathrm{DNN}\big(\mathcal{I}(t)\big), \label{algo:dnnpso_next_position_norm} \\
  \Pos(t+1) &= E_{{\it bound}} \cdot \hat{\Pos}(t+1), \label{algo:dnnpso_next_position}
\end{align}
where $\mathcal{I}(t)$ denotes the input features provided to the network at $t$. Our implementation of the DNN outputs a normalized coordinate $\hat{\Pos}(t+1)$ bounded within $[-1,1]$. The actual next position $\Pos(t+1)$ is obtained by scaling this normalized coordinate by $E_{{\it bound}}$ and restoring it to the coordinate space of the environment.
\par

\subsubsection{Observation Ring Radius Multiplier $(m_i)$}
Each particle $p_i$ has a radius multiplier $m_i$ that controls its observation rings with multiple radii (Fig.~\ref{fig:dnnpso_particle}). The radius multiplier enables the particles to balance exploration and exploitation. When a particle fails to improve the utility, the multiplier expands its observation area for broader exploration; however, after improvement, it contracts the area to focus on promising regions.
\par

The multiplier, initialized as $m_i = 1$, serves as a scaling factor $m_{\mathrm{fac}}$ to adjust its value dynamically. A change counter $\gamma_i$ prevents the radii from becoming excessively large, which would cause sparse or small observations and limit particle exploration. The scaling factor $m_{\mathrm{fac}} \in (0,1)$ affects particle performance; larger values cause drastic radius changes that risk missing global optima, whereas smaller values make particles too slow to react to changes.
\par

If a particle moves to a position with a higher utility value $\util(\Pos_i(t),t) < \util(\Pos_i(t+1),t+1)$, the observation radius is decreased by the scaling factor:
\begin{equation} \label{eqn:dnnpso_shrink_ring}
m_i \leftarrow m_i \cdot m_{\mathrm{fac}}, \quad \gamma_i \leftarrow \gamma_i - 1
\end{equation}
Otherwise, if the move results in a lower utility value,
\begin{equation} \label{eqn:dnnpso_expand_ring}
m_i \leftarrow \frac{m_i}{m_{\mathrm{fac}}}, \quad \gamma_i \leftarrow \gamma_i + 1
\end{equation}
To maintain exploration-exploitation stability, $\gamma_i$ is bounded within $[\gamma_{\min}, \gamma_{\max}]$. When $\gamma_i = \gamma_{\max}$, the particle is stagnant and cannot explore effectively. The particle then undergoes re-randomization, as described in Section~\ref{section:re_random}.
\par

\subsubsection{Re-randomization}\label{section:re_random}
When a particle $p_i$ becomes stagnant, its position $\Pos_i(t)$ is re-randomized to escape the local optimum and explore other regions. Because the environment is dynamic, the stored global best position $\GlobalBest(t)$ may become outdated. During re-randomization, the particle tests if $\GlobalBest(t)$ remains unchanged from the previous state. If the utility value of   $\GlobalBest(t)$ at the current state $\util(\GlobalBest(t),t)$ has changed compared to the previous state $\util(\GlobalBest(t),t-1)$, the particle updates the stored $\util(\GlobalBest(t),t)$. A random Boolean determines whether the particle moves to the current global best position or a random coordinate.
\par

After re-randomization, the particle updates the new position utility value. The observation point set $Q_i$ is recomputed, and both the ring multiplier $m_i$ and counter $\gamma_i$ are reset. The re-randomization algorithm for particle $p_i$ is defined in Algorithm~\ref{algo:dnnpso_re_random}, where $\mathcal{U}$ denotes a uniform distribution.
\par

\begin{algorithm}[t]
\caption{\textsc{Re-randomization}($p_i$)}\label{algo:dnnpso_re_random}
\begin{algorithmic}[1]
  \STATE // For particle $i\in\Particles$; $b\in\{0,1\}$ selected randomly
  \IF{$b=1$  // with probability $0.5$}
    \STATE $\Pos_i(t+1) \leftarrow \GlobalBest(t)$
    \IF{$\util(\GlobalBest(t), t-1) \neq \util(\GlobalBest(t),t)$} 
\STATE Update $\util(\GlobalBest(t),t)$;
\ENDIF
\ELSE
    \STATE $\Pos_i(t+1) \sim \mathcal{U}\big([-E_{{\it bound}}, E_{{\it bound}}]\big)$
\ENDIF
\STATE Update $\util(\Pos(t+1),t)$;
\STATE Update $Q_i(t)$ using Eq.~\ref{eqn:dnnpso_orbit_update};
\STATE $m_i \leftarrow 1$, $\gamma_i \leftarrow 0$ // Reset ring multiplier and counter
\end{algorithmic}
\end{algorithm}

\subsubsection{Deep Neural Network Structure}
Each predictive network is implemented as a fully connected {\em   multi-layer perceptron} (MLP). The input and hidden layers use a {\em rectified linear unit} (ReLU) activation function, whereas the output layer uses hyperbolic tangent ($\tanh$) activation to produce a normalized next-coordinate prediction $\hat{\Pos}(t+1) \in [-1,   1]$. This ensures that the predicted coordinates remain within the normalized environmental bounds.
\par

To enhance the convergence stability with ReLU activation, the network weights use He initialization~\cite{He2015DelvingDI}, whereas the biases are set to zero. The input layer size is $|\mathcal{I}(t)|$, and the output layer dimensions match the two-dimensional coordinate space. The hidden architecture comprises three fully connected layers, each containing 16 neurons. For angular and radial features $\bm{\theta}_i$ and $\bm{r}_i$ in our experiment, this MLP configuration yields the lowest average tracking error $\mu_{\mathcal{E}_i} \pm \sigma_{\mathcal{E}_i}$ across environmental groups.
\par

\subsection{Pre-training of Networks} \label{sec:pretraining}
Before running the full experiments on our proposed method, we pre-trained each network in a simplified version of the target dynamic environment. We found that pre-training improves convergence performance and stability by allowing the network to learn general spatial and temporal patterns prior to online adaptation.
\par

For pre-training, we generate environments using the parameters listed in Table~\ref{tbl:env_experiment_param_list} with several   changes. Each network is trained independently, and no pre-training environment is reused across networks to prevent overfitting to specific environmental dynamics.
\par

During pre-training, each particle is simulated from $t = 1$ to $t_{\max}$, constructing the network input set $\mathcal{I}(t)$ (Algorithm~\ref{algo:dnnpso_inputset}) and updating positions via Algorithms~\ref{algo:dnnpso_next_position_norm} and~\ref{algo:dnnpso_next_position}, with the expected coordinate $\ActualGB(t)$ used as the target label for pre-training.
\par

As shown on Fig.~\ref{fig:peaks_global}, the position of $\ActualGB(t)$ cannot be predicted relative to any peak $h_i$. Therefore, $\ActualGB(t)$ is obtained by treating each state of the environment as static environment. For each state at $t$, we identified $\ActualGB(t)$ by using canonical PSO with swarm size five times larger than the peak numbers in the environment ($|p| = 5 \cdot |H_{i,j}|$) and iterated until the global best position $\GlobalBest(t)$ shows no improvement for five consecutive steps, and repeated for 3 times to minimize the probability of misidentifying $\ActualGB(t)$. For this process, we used velocity weight $w = 0.1$ and coefficients $c_1 = c_2 = 2$.
\par

The expected coordinates $\ActualGB(t)$ are normalized to $\hat{a}(t)$ and paired with the network inputs $\mathcal{I}(t)$ to form the training dataset. The dataset is randomly sampled into mini-batches of size $\tau_{\mathrm{size}}$ for a stable gradient estimation during optimization.
\par

The network is trained using the {\em mean squared error} (MSE) loss function and optimized using AdaGrad~\cite{Duchi2011AdaptiveSM}. MSE loss was chosen because it provides a differentiable measure of the Euclidean deviation between the predicted and expected coordinates, directly aligning with the minimization of the positional error. Its quadratic form penalizes larger deviations more heavily, thereby encouraging precision in dynamic environments.
\par

The AdaGrad optimizer was chosen for its ability to adapt each parameter's learning rate individually, enabling faster convergence with sparse or unevenly scaled gradients. To improve convergence stability, a cosine annealing learning rate scheduler was employed, which reduced the rate smoothly and helped the model escape shallow local minima during the later stages of training. A warm-up phase was introduced, where the learning rate increased linearly from $\alpha_{\mathrm{start}}$ to $\alpha_{\mathrm{warm}}$ during the first batch before following a cosine decay to $\alpha_{\mathrm{final}}$. This warm-up strategy mitigates training instability during initialization by preventing abrupt updates before the optimizer's adaptive statistics are established.
\par

Each network was trained for $\tau_{\mathrm{epoch}}$ epochs using $(\mathcal{I}(t), \hat{a}(t))$. The pre-training process was repeated three times with different random particle positions to enhance generalization.
\par

\begin{algorithm}[tb]
\caption{Algorithm Overview}\label{algo:dnnpso_main}
\begin{algorithmic}[1]
\STATE $G \leftarrow []$; // Global best history
\STATE // Initialize swarms
\FOR{$\forall p_i \in \mathcal{P}$}
\STATE $\Pos_i \leftarrow \mathcal{U}(-E_{bound}, E_{bound})$; // Randomly initialize position
\STATE Calculate $\util(\Pos_i,t)$;
\STATE Update $Q_i$ using Eq.~\ref{eqn:dnnpso_orbit_update};
\IF{\dnngpso} \label{algo:dnnpso_main:mdnnpso_training_data_init}
\STATE $D_i(t) \leftarrow []$; // Received training data
\ENDIF
\ENDFOR
\STATE Load pre-trained model(s);
\WHILE{$t < t_{max}$ // Until terminal condition is reached}
\STATE Execute Function {\sc updateSwarm()}; \label{algo:dnnpso_main:update_swarm}
\STATE $G \leftarrow G \cup \GlobalBest(t)$; // Append global best to history
\STATE $\forall h_{i,j,k} \in H_{i_j}, h_{i,j,k} \leftarrow h_{i,j,k} + \boldsymbol{\epsilon}_{i,j,k}(t) $; // Move peaks
\STATE  $t \leftarrow t + 1$; // Progress time
\ENDWHILE
\STATE Return $G$;
\end{algorithmic}
\end{algorithm}

\subsection{Neural-Network-Guided PSO (\nngpso)}
We introduce two variants of the \nngpso algorithm with identical particle configurations, as shown in Fig.~\ref{fig:dnnpso_particle}. Both variants train networks upon successful exploration to adapt the swarms to the environment, thereby improving the convergence performance. The differences between the variants are the neural network scope across the swarm and the information propagation method used.
\par

Swarm initialization begins by generating particles. Each particle is assigned a random position within the environment bounds, followed by the computation of its observation points $Q_i$. For the distributed \nngpso (\dnngpso), as described below, an array $D_i$ is initialized to record the network training data propagated from other particles at each time $t$ (Line~\ref{algo:dnnpso_main:mdnnpso_training_data_init}).
\par

At time $t$, the swarm is updated (Line~\ref{algo:dnnpso_main:update_swarm}) using Algorithm~\ref{algo:dnnpso_update} or Algorithm~\ref{algo:mdnnpso_update}, depending on the variant. The current global best position $\GlobalBest(t)$ is added to the history list, and each peak $h_{i,j,k}$ moves towards its assigned center $c_{i,j,k}$ with a random velocity $\boldsymbol{\epsilon}_{i,j,k}(t)$. When $t = t_{\max}$, the global best history $G$ is used to compute the cumulative tracking error, as defined in Eq. ~(\ref{eqn:tracking_error}).
\par

\subsubsection{Centralized Neural Network-guided PSO (\cnngpso)} 
The first variant uses a single neural network that is shared by all the swarm particles. Unlike in canonical PSO, the particles communicate indirectly through the shared neural network. For every successful exploration, the particles train the shared neural network using information that improves their positions.
\par

As shown in Algorithm~\ref{algo:dnnpso_update}, at time $t$, the swarm updates the observation point locations $Q_i$ of particle $\forall i\in\Particles$ before calculating the next position $\Pos_i(t+1)$. After the particles move to their next positions, they adjust the ring multipliers, $m_i$. If next positions have better utility values, particles train the central network using input set $\mathcal{I}_i(t)$ for $\forall i$ and normalized next position $\hat{\Pos}_i(t+1)$ as the label at epoch $\tau_{\mathrm{epoch}}^c$ ($\tau_{\mathrm{epoch}}^c>0$). When a particle reaches the ring adjustment counter limit $\gamma_i = \gamma_{max}$, it is randomly re-initialized. Finally, the global best position is updated.
\par

\begin{algorithm}[tb]
\caption{\cnngpso \xspace {\sc updateSwarm()}}\label{algo:dnnpso_update}
\begin{algorithmic}[1]
\FOR{$\forall p_i \in \mathcal{P}$}
\STATE Update $Q_i$ using Eq.~\ref{eqn:dnnpso_orbit_update};
\STATE Generate $\mathcal{I}_i(t)$ using Eq.~\ref{algo:dnnpso_inputset};
\STATE Update $\Pos_i(t+1)$ using Eq.~\ref{algo:dnnpso_next_position_norm} and Eqn.~\ref{algo:dnnpso_next_position};
\IF{$\util(\Pos_i(t+1),t) > \util(\Pos_i(t),t)$}
\STATE Update $m_i$ and $\gamma_i$ using Eq.~\ref{eqn:dnnpso_shrink_ring}
\STATE Train network with $\mathcal{I}_i(t)$ and $\hat{x}_i(t+1)$ with $\tau_{\mathrm{epoch}}^c$ epoch;
\ELSIF{$\util(\Pos_i(t+1),t) < \util(\Pos_i(t),t)$}
\STATE Update $m_i$ and $\gamma_i$ using Eq.~\ref{eqn:dnnpso_expand_ring};
\ENDIF
\IF{$\gamma_i = \gamma_{max}$}
\STATE Re-randomize $p_i$ using Alg.~\ref{algo:dnnpso_re_random};
\ENDIF
\IF{$\util(\GlobalBest(t),t) < \util(\Pos_i(t+1), t)$}
\STATE $\GlobalBest(t) \leftarrow \Pos_i(t+1)$; // Update global best
\ENDIF
\ENDFOR
\end{algorithmic}
\end{algorithm}

\subsubsection{Distributed Neural Network-guided PSO (\dnngpso)} 
The second algorithm is a decentralized version of \dnngpso, where each particle has its own neural network. Because particles are initialized randomly, with networks trained independently and the training data coming from other particles, a particle would not be too specialized in its vicinity owing to overtraining using self-generated data.
\par

The \dnngpso algorithm shown in Algorithm~\ref{algo:mdnnpso_update} is similar to \cnngpso (Algorithm~\ref{algo:dnnpso_update}), with the main difference being the learning network structures.  Upon successful exploration, the particle propagates the network input set $\mathcal{I}_i(t)$, normalized output $\hat{\Pos}_i(t+1)$, and current time step $t'$ (Line~\ref{algo:mdnnpso_update:propagate}). The data received from the previous time step $D_i(t-1)$ are included, as the particle can improve its position from the training data received from other particles at time step $t-1$. 
\par

After the particles move to their next positions, they train networks using the recorded training data in $D_i(t)$. Unlike \cnngpso, the training epoch in \dnngpso is determined by the timestep in each tuple $d_j$. The epoch would be $\tau_{\mathrm{epoch}}^d$ if data are received in the current time step $t' = t$, whereas it would be $\tau_{\mathrm{epoch}}^d - 1$ if received in the previous time step. This varying epoch trains the networks less on the old training data. Because particles may receive data generated by themselves through other particles, training data from the same particle from the previous time step are skipped to prevent over-training.
\par

\begin{algorithm}[tb]
\caption{\dnngpso \xspace {\sc updateSwarm()}}\label{algo:mdnnpso_update}
\begin{algorithmic}[1]
\FOR{$\forall p_i \in \mathcal{P}$}
\STATE Update $Q_i$ using Eq.~\ref{eqn:dnnpso_orbit_update};
\STATE Generate $\mathcal{I}_i(t)$ using Eq.~\ref{algo:dnnpso_inputset};
\STATE Update $\Pos_i(t+1)$ using Eqs.~\ref{algo:dnnpso_next_position_norm} and ~\ref{algo:dnnpso_next_position};
\IF{$\util(\Pos_i(t+1),t) > \util(\Pos_i(t),t)$}
\STATE Update $m_i$ and $\gamma_i$ using Eq.~\ref{eqn:dnnpso_shrink_ring};
\STATE // Propagate information
\FOR{$\forall p_j \in \mathcal{P}, i \neq j$ } \label{algo:mdnnpso_update:propagate}
\STATE $D_j(t) \leftarrow D_j(t)\cup \{\big(\mathcal{I}_i(t), \hat{x}_i(t+1), t'\big)\} \cup D_i(t-1)$;
\ENDFOR
\ELSIF{$\util(\Pos_i(t+1),t) < \util(\Pos_i(t),t)$}
\STATE Update $m_i$ and $\gamma_i$ using Eq.~\ref{eqn:dnnpso_expand_ring};
\ENDIF
\IF{$\gamma_i = \gamma_{max}$}
\STATE Re-randomize $p_i$ using Alg.~\ref{algo:dnnpso_re_random};
\ENDIF
\IF{$\util(\GlobalBest(t),t) < \util(\Pos_i(t+1),t)$}
\STATE $\GlobalBest(t) \leftarrow \Pos_i(t+1)$; // Update global best
\ENDIF
\ENDFOR
\STATE // Train networks
\FOR{$\forall p_i \in \mathcal{P}$}
\FOR{$\forall d_j \in D_i(t), d_j = \{\big(\mathcal{I}_i(t), \hat{x}_i(t+1), t'\big)\}, i \neq j$} 
\IF{$\boldsymbol{\kappa}=\tau_{\mathrm{epoch}}^d -(t-t'), \boldsymbol{\kappa} > 0$}
\STATE Train network using $d_j$ with $\boldsymbol{\kappa}$ epoch;
\ENDIF
\ENDFOR
\ENDFOR
\end{algorithmic}
\end{algorithm}

\section{Experimental Setup}
\subsection{Environment Setting}\label{sect:cnngpso_env_setting}
We define a collection of environment groups, denoted by $\mathcal{E}_i$, where all environments in the group share the same environment parameters. Each group $\mathcal{E}_i$ contains multiple environments $E_{i,j}$, such that $E_{i,j} \in \mathcal{E}_i$ represents the $j$th environment in the $i$th group. An environment $E_{i,j}$ consists of two sets: a set of peaks $H_{i,j} = \{h_{i,j,1}, h_{i,j,2}, \ldots\}$ and a set of centers $C_{i,j} = \{c_{i,j,1}, c_{i,j,2}, \ldots\}$.
\par

All environments share spatial boundaries defined by a square search space with dimensions having bounds $[-E_{\mathrm{bound}},\,   E_{\mathrm{bound}}]$, whereas peak and center initial positions are sampled within a smaller area $[-E'_{\mathrm{bound}},\,   E'_{\mathrm{bound}}]$, where $E'_{\mathrm{bound}} = E_{\mathrm{factor}} \cdot E_{\mathrm{bound}}$. Constraining the initialization to a smaller   region helps maintain swarm diversity during early iterations, which   enhances the exploration performance~\cite{source1}.
\par

Each peak $h_{i,j,k}$ is generated as a bivariate Gaussian distribution with a mean and SD of
\[
\mu \sim \mathcal{U}(-E'_{{\it bound}},\, E'_{{\it bound}}), \sigma \sim \mathcal{U}(\sigma_{\min},\, \sigma_{\max}),
\]
by assuming dimensional independence (i.e., $\rho = 0$). To ensure smoother utility landscapes, we imposed bounds $\sigma_{\min}$ and $\sigma_{\max}$ on the SD to prevent peaks from becoming too concentrated and dominating utility values while ensuring that the peak has a non-negligible effect on $\util(\Pos,t)$.
\par

Each center $c_{i,j,k}$ is placed randomly in the environment according to $c_{\mu} \sim \mathcal{U}\big( -E'_{{\it bound}},\, E'_{{\it bound}})$, and peak $h_{i,j,k}$ moves linearly towards its assigned center $c$ with a random velocity drawn from
\begin{equation}
    \boldsymbol{\epsilon}_{i,j,k}(t) \sim \mathcal{U}\Big(0,\, \frac{||c-h_{i,j,k}(0)||_2}{\omega}\Big),
\end{equation}
until it reaches its destination. To maintain a dynamic environment, we set $\omega = t_{\max}$, preventing peaks $H_{i,j}$ from moving so slowly that the landscape becomes quasi-static or so quickly that all peaks reach their centers and the environment changes. For each environment group $\mathcal{E}_i$, we generated $E_{\it   count}$ environments, and each environment $E_{i,j}$ was evaluated through $E_{\it run}$ runs with the swarm population reinitialized randomly before each run for statistical robustness.
\par

\begin{table}
\caption{List of experimental parameters}
\centering
\begin{tabular}{lll}
\toprule
Symbol & Description & Value  \\
\midrule
$E_{\it bound}$ & $E$ bounds for each dimension & $10$ \\
$E_{\it factor}$ & Sampling area factor for $E$  & $0.8$ \\
$E_{\it count}$ & Number of environments per $\mathcal{E}_i$ & $5$ \\
$E_{\it run}$ & Number of experiments per $E$ & $3$ \\
$\sigma_{\min}$ & Lower bound of $\sigma$ for $H$ & $0.25$ \\ 
$\sigma_{\max}$ & Upper bound of $\sigma$ for $H$ & $1$ \\
$t_{\max}$ & Maximum experiment time & $20,000$ \\
$\omega$ & Peak movement speed factor & $20,000$ \\
\bottomrule
\end{tabular}
\label{tbl:env_experiment_param_list}
\end{table}

\begin{table}
\caption{Parameter settings for each environment group $\mathcal{E}_i$.}
\centering
\begin{tabular}{lll}
\toprule
Env. group (\textbf{$\mathcal{E}_i$}) & No. of peaks
($|H|$) & No. of centers ($|C|$)  \\
\midrule
$\mathcal{E}_1$ & 100 & 50  \\ 
$\mathcal{E}_2$ & 200 & 100  \\
$\mathcal{E}_3$ & 200 & 200  \\
$\mathcal{E}_4$ & 500 & 100  \\ 
\bottomrule
\end{tabular}
\label{tbl:env_setting}
\end{table}

\subsection{Measurement Method} \label{sect:measurement}
To evaluate the effectiveness of the proposed and comparison methods in tracking the global optimum, we measured the cumulative Euclidean distance between the true global optimum $\ActualGB(t)$ and the position identified by the swarm $\GlobalBest(t)$ during experimental runs in environment $E_{i,j}$. The details of finding $\ActualGB(t)$ are provided in Section~\ref{sec:pretraining}. Formally, the tracking error for a single run is defined as
\begin{equation} \label{eqn:tracking_error}
    \mathrm{err}(E_{i,j}^k) = \sum_{t=1}^{t_{\max}} \lambda(t) \cdot ||\GlobalBest(t) -\ActualGB(t)||_2,
\end{equation}
where $\GlobalBest(t)$ denotes the global optimum position detected by the swarm, and $\ActualGB(t)$ denotes the actual global optimum position in the environment at time $t$. The weighting factor $\lambda(t)$ at timestep $t$ is given by
\[
\lambda(t) =
\begin{cases}
1, & \text{if } \util(\GlobalBest(t),t) \leq \util(\ActualGB(t),t),\\
1.5, & \text{otherwise.}
\end{cases} \label{eqn:penalty}
\]
The additional penalty ($\lambda=1.5$) increases the error when the swarm fails to recognize a decrease in the true global optimum reward value, indicating convergence around an outdated region, whereas the actual optimum has shifted elsewhere.
\par

For each environment $E_{i,j}$, the mean and SD of tracking error across $k_{\max}>0$ experimental runs are computed as
\begin{align}
    \mu_{E_{i,j}} &= \frac{1}{E_{\it run}} \sum_{k=1}^{E_{\it run}} \mathrm{err}(E_{i,j}^k),\\
    \sigma_{E_{i,j}} &= \sqrt{\frac{1}{E_{\it run}} \sum_{k=1}^{E_{\it run}}
    \big(\mathrm{err}(E_{i,j}^k) - \mu_{E_{i,j}}\big)^2}.
\end{align}
We then aggregate the results for all environments within each environment group $\mathcal{E}_i$ as
\begin{align} \label{eqn:aggregate_mean}
    \mu_{\mathcal{E}_i} &= \frac{1}{E_{\it count}} \sum_{j=1}^{E_{\it count}} \mu_{E_{i,j}},\\
\label{eqn:aggregate_sd}
    \sigma_{\mathcal{E}_i} &= \sqrt{\frac{1}{E_{\it count}} \sum_{j=1}^{E_{\it count}}
    \big((\sigma_{E_{i,j}})^2 + (\mu_{E_{i,j}} - \mu_{\mathcal{E}_i})^2\big)}.
\end{align} 
Finally, the tracking performance of each algorithm in the environment group $\mathcal{E}_i$ is reported as $\mu_{\mathcal{E}_i} \pm \sigma_{\mathcal{E}_i}$, showing the average tracking error and variability across the experiments. Lower error values indicate a better global optimum position tracking.
\par

\subsection{Comparison Algorithms}
We compared our method with canonical PSO and two recent PSO variants~\cite{PSPSO,SPSO} that incorporate speciation and subswarm mechanisms for dynamic optimization as baseline methods. For PSPSO and SPSO, we used hyperparameters recommended by their authors, except for species or subswarm size and count, which were adjusted to ensure equivalent total observations to our proposed methods. Because our method uses $|\Particles| = 5$ particles with $1 + |Q| = 33$ observation points per particle (165 in total), all the comparison algorithms were configured with matching particle distributions to ensure fairness.
\par

\begin{table}
\caption{Parameter settings for \nngpso variants.}
\centering
\begin{tabular}{ll}
\toprule
Parameter & Value  \\
\midrule
$m_\mathrm{fac}$ & 0.8 \\
$\gamma_{\min}$  & -10 \\
$\gamma_{\max}$ & 10 \\
$\bm{\theta}_i$ & $0, \frac{\pi}{4}, \frac{\pi}{2}, \frac{3\pi}{4},
\pi, \frac{5\pi}{4}, \frac{3\pi}{2}, \textrm{ or }\frac{7\pi}{4}$ \\
$\bm{r}_i$ & $0.5, 1.0, 1.5, \textrm{ or } 2.0$\\
$\tau_{\mathrm{epoch}}^{c}$ & 1\\
$\tau_{\mathrm{epoch}}^d$ & 2 \\
\bottomrule
\end{tabular}
\label{tbl:nngpso_setting}
\end{table}

The parameters for both \nngpso variants are listed in Table~\ref{tbl:nngpso_setting}. For network pre-training, we used the algorithm parameters (Table~\ref{tbl:env_experiment_param_list}) with peaks $|H| \in \{25,50\}$, centers $|C| \in \{5,10\}$, $t_{\max}=10,000$, batch size $\tau_{\mathrm{size}} = 64$, $\tau_{\mathrm{epoch}}=5$ epochs, starting learning rate $\alpha_{\mathrm{start}} = 0$, warmed up learning rate $\alpha_{\mathrm{warm}} = 10^{-3}$, and final learning rate $\alpha_{\mathrm{final}} = 10^{-5}$.
\par

\subsubsection{Canonical PSO}
We include the canonical PSO algorithm, as described in Section~\ref{sect:cpso}. The cognitive and social coefficients were set to $c_1 = c_2 = 2$, and velocity weight was set to $w = 0.1$. To ensure consistency with our proposed methods, we conducted experiments using particle counts of $|p| \in \{5, 165\}$.
\par

\subsubsection{Perturbation and Speciation-based PSO}
The {\em perturbation and speciation-based PSO } (PSPSO)~\cite{PSPSO} is a multi-swarm PSO for dynamic environments that operates without change detectors. The population is split into independent subpopulations via speciation to track multiple regions in parallel. PSPSO performs standard PSO updates within the active subpopulations. To prevent stagnation, it removes the subpopulation with the worst local best position when the two merge within a threshold. The converted subpopulations are then deactivated. Diversity is maintained by perturbing the particles in a random subpopulation. When active particles fall below a threshold, it archives the best particle from each deactivated subpopulation to create new particles using archived particles as seeds. This process maintains exploration and exploitation, enabling the tracking of moving optima without explicit change detection. In our experiment, we used two variants with subpopulation sizes of $5$ and $33$.
\par

\subsubsection{Species-based PSO with Adaptive Population and Deactivation}
The {\em species-based PSO with adaptive population and deactivation} ($\mathrm{SPSO}_{\mathrm{+AP+AD}}$)~\cite{SPSO}, or SPSO for short, maintains the swarm as a species, similar to PSPSO. Given a species size $m$, the best unassigned particles become species seeds with $m-1$ nearest particles assigned to them, with seeds serving as the species' best. A species is labeled as a tracker when its spatial size falls below the tracking threshold. To avoid redundancy, an exclusion rule removes the weaker of two species whose seeds are closer than the preset exclusion distance. A deactivation radius $r_a$ controls tracker deactivation, whereas species with the global best remain active to track the potential global best. Particle diversity is injected when all species converge within $r_{\mathrm{generate}}$ by injecting $m$ new individuals or re-randomizing the worst species. Upon environmental change, SPSO re-diversifies tracker species by scattering non-seed members using the estimated shift severity $\hat{s}^{(t)}$, re-evaluates the fitness of all particles, and resets deactivation controls in proportion to $\hat{s}^{(t)}$. For our experiment, we used species sizes of $n \in \{5,33\}$.
\par

\begin{table*}
\caption{Tracking error comparison against baselines}
\centering
\begin{tabular}{llllllll}
\toprule
\textbf{Algorithm} & \textbf{$|p|$} & \textbf{$n$} &\multicolumn{4}{c}{\textbf{$\mathrm{error}(\mathcal{E}_i)$ per Environment Group}} &\textbf{$\mathrm{error}(\bar{\mathcal{E}})$} \\
& & & $\mathcal{E}_1$ & $\mathcal{E}_2$ & $\mathcal{E}_3$ & $\mathcal{E}_4$ & \\
\midrule
\cnngpso  & 5   & - & $5.331 \pm 4.464$ & $5.827 \pm 3.603$ & $5.350 \pm 3.753$ & $5.471 \pm 3.747$ & $5.495 \pm 3.911$ \\
\dnngpso & 5   & - & \textbf{4.552 $\pm$ 4.370} & \textbf{5.213 $\pm$ 3.879} & \textbf{4.786 $\pm$ 3.611} & \textbf{3.858 $\pm$ 3.798}  & \textbf{4.602 $\pm$ 3.955} \\
PSO     & 5   & - & $8.595 \pm 4.347$&$10.196 \pm 4.576$ & $8.526 \pm 4.042$ & $7.995 \pm 3.926$ & $8.828 \pm 4.310$ \\
PSO     & 165 & - & $8.804 \pm 4.893$& $9.486 \pm 4.773$ & $8.356 \pm 4.092$ & $8.583 \pm 4.242$  & $8.818 \pm 4.423$ \\
PSPSO   & 165 & 5 & $16.889 \pm 7.060$&$18.878 \pm 8.310$ & $18.085 \pm 8.266$ & $16.739 \pm 8.919$  & $18.503 \pm 7.507$ \\
PSPSO   & 165 & 33 & $18.211 \pm 5.619$& $20.198 \pm 6.919$ & $19.904 \pm 6.694$ & $19.118 \pm 6.965$ & $19.358 \pm 6.617$ \\
SPSO    & 165 & 5 & $5.265 \pm 5.998$&$6.576 \pm 7.149$ & $6.442 \pm 6.890$ & $4.932 \pm 6.414$  & $5.804 \pm 6.666$ \\
SPSO    & 165 & 33 &  $5.641 \pm 6.457$&$6.431 \pm 7.151$ & $6.492 \pm 6.965$ & $5.625 \pm 6.729$ & $5.926 \pm 6.756$ \\
\bottomrule
\end{tabular}
\label{tbl:dnnpso_algo_comparison}
\end{table*}

\section{Results and Discussion}\label{sect:exp_result}
\subsection{Comparison with Baselines}
We compared our proposed methods with the baseline algorithms under identical experimental parameters (Table~\ref{tbl:env_experiment_param_list}) and observation counts using different environmental groups (Table~\ref{tbl:env_setting}). Each generated environment used the same random seed for identical peak and center positions and movements. The comparison focused on the mean and SD error values obtained from the experiments within each environment group $\mathcal{E}_i$ using the measurement methods defined in Section~\ref{sect:measurement}.
\par

Table~\ref{tbl:dnnpso_algo_comparison} lists the quantitative results, including the algorithm name, particle count $|p|$, subspecies count $n$, and error values for each environment group, with aggregated values $\bar{\mathcal{E}}$ across all groups computed using Eqs.~\ref{eqn:aggregate_mean} and~\ref{eqn:aggregate_sd}. For the canonical PSO and our proposed method, the species count $n$ was left blank.
\par

Comparing the aggregate error values $\bar{\mathcal{E}}$ across environmental groups, both variants  of our method(\cnngpso and \dnngpso) produced the lowest error mean and standard deviation compared to the baselines. The neural network component appeared to learn the environment behavior allowing particles to more reliably predict the next the global optimum through observation points, improving convergence performance and adapting capability to track the global optimum position throughout the experiments.  \dnngpso showed better error values with 16.25\% lower mean but 1.11\% higher SD than \cnngpso. This advantage can be attributed to the neural networks in \dnngpso initialized with separate random weights and pre-training, improving particle diversity and thus the swarm exploration capability.
\par

\begin{figure}
  \centering
  \begin{subfigure}[t]{0.7\linewidth}
      \includegraphics[width=\linewidth]{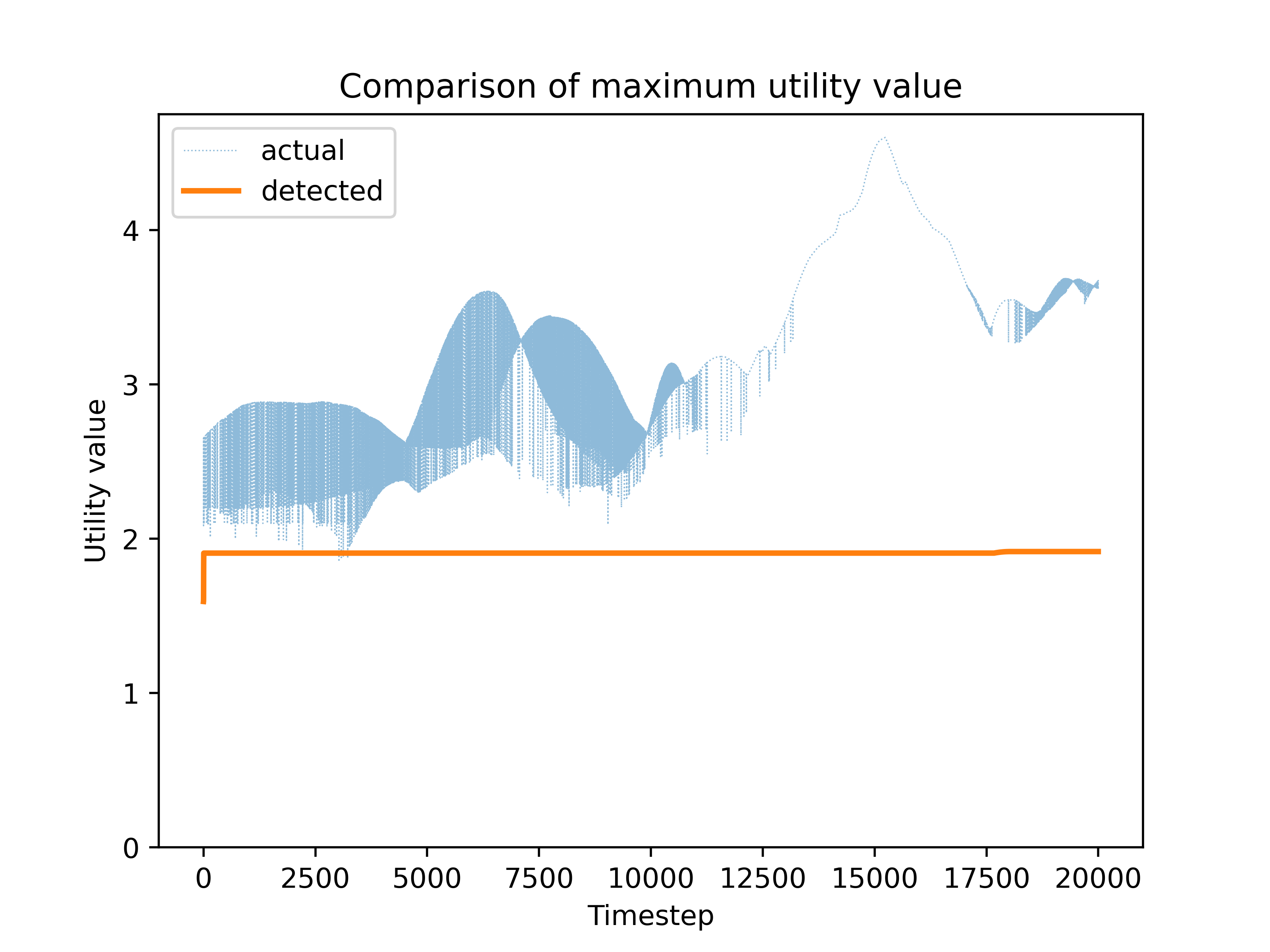}
      \caption{Canonical PSO with $|p| = 165$}
            \Description[]{A figure of an experiment run of canonical PSO in an environment with 165 particles. While the utility value of actual global optimal increases and decreases as the time progresses, PSO quickly converged into a local optima and unable to escape from it, causing PSO to unable to track the global optimal even when the utility value of global best is increasing.}

      \label{fig:pso_trapped}
  \end{subfigure}
  \begin{subfigure}[t]{0.7\linewidth}
      \includegraphics[width=\linewidth]{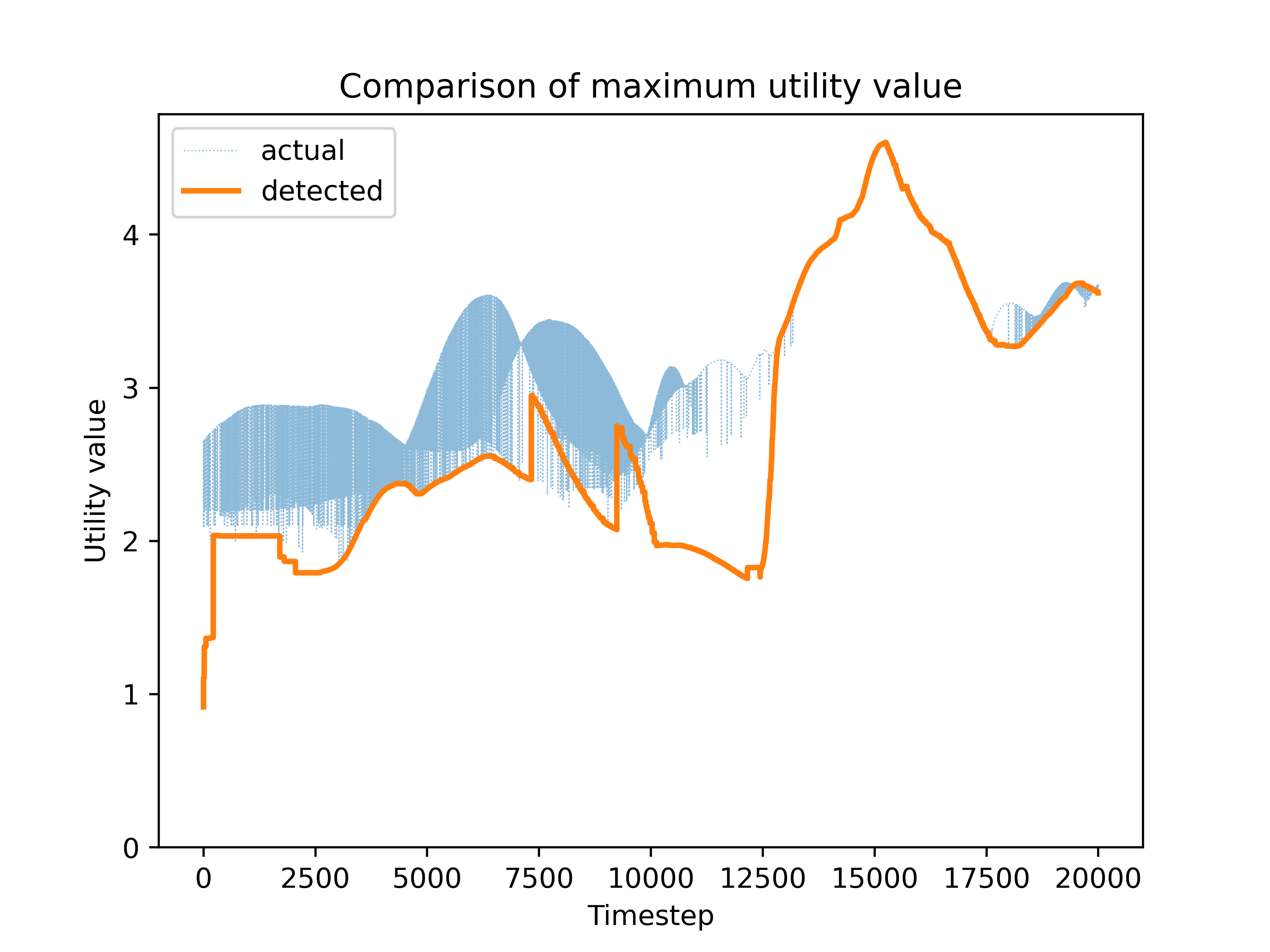}
      \caption{\dnngpso with $|p| = 5$}
            \Description[]{A figure of an experiment run of DNNGPSO variant
        of our proposed method. While the utility value of actual
        global optimal increases and decreases as the time progresses,
        in this experiment run DNNGPSO particle was able to identify
        the global best early, and able to track the global best
        throughout the experiment. However, in periods when utility
        value of global best was rapidly changing, DNNGPSO tracked the
        lower bound.}
      \label{fig:dnn_tracking}
  \end{subfigure}
  \caption{Utility values of actual global best position $\ActualGB(t)$ and detected
        global best position $\GlobalBest(t)$ in $E_{3,2}^1$}
    \label{fig:experiment_samples}
\end{figure}

Figure~\ref{fig:experiment_samples} shows the maximum utility values of the actual global best position $\ActualGB(t)$ and detected global best position $\GlobalBest(t)$ in $E_{3,2}^1$ at each time $t$. When the peaks in $H$ have not converged to the assigned centers and remain dispersed throughout the environment, $\util(\ActualGB(t),t)$ may change dramatically at every $t$ as new global optimal points appear or the existing ones disappear. This rapid fluctuation of $\util(\ActualGB(t),t)$ appears in Figure~\ref{fig:experiment_samples} for example, at $0 \leq t \leq 10000$, where the dotted blue line appears as a blue area owing to frequent fluctuations. In our experiments, all the algorithms had difficulty identifying and tracking $\ActualGB(t)$  during this rapid fluctuation period.
\par

Figure~\ref{fig:pso_trapped} shows the canonical PSO tendency to converge prematurely to local optima, even with $|\Particles|=165$. The swarm remained trapped after converging to the local optima, even when $\util(\ActualGB(t),t)$ increased. Without the adaptation capability, the canonical PSO swarm cannot adapt to decreasing $\util(\ActualGB(t),t)$ nor can it escape from a local optimum.
\par

In contrast, Fig.~\ref{fig:dnn_tracking} shows that \dnngpso tracked the global best position when $\util(\ActualGB(t),t)$ increased or decreased throughout the experiment. Table~\ref{tbl:dnnpso_algo_comparison} shows that \dnngpso achieved an average error reduction of 47.81\% across environments when comparing $\mathrm{error}(\bar{\mathcal{E}})$ between canonical PSO and \dnngpso.
\par

However, the network in our method uses the MSE to minimize the distance between $\GlobalBest(t)$ and $\ActualGB(t)$. This allowed the \nngpso swarm to track $\GlobalBest(t)$ with minimal error when $\util(\GlobalBest(t),t)$ changed gradually, as shown in Fig.~\ref{fig:dnn_tracking}. However, because the distance is an error function, swarms tend to be conservative when $\util(\GlobalBest(t),t)$ fluctuates rapidly, tracking the lower fluctuation bound instead of the maximum, such as when $0 \leq t \leq 10000$ in Fig.~\ref{fig:dnn_tracking}. Furthermore, \dnngpso is not completely immune to misidentifying a local optimal as the global optimal, such as when $10000 \leq t \leq 12500$. 
\par

Despite employing a subpopulation with re-diversification, PSPSO yields the largest average cumulative tracking error among all methods, including our proposed methods, for both configurations $n \in \{5,33\}$. In model environments, the peak motion can place the global optimum at time $t$ far from that at $t-1$. As shown in Fig.~\ref{fig:experiment_samples}, which shows the utility values of the actual global best position $\ActualGB(t)$ and the detected global best position $\GlobalBest(t)$ in $E_{3,2}$, the utility of the global optimum may decrease while the peaks have not converged to their assigned centers. Although PSPSO can rapidly converge to the global optimum, the subsequent optimum may be far away with a lower utility, leading to frequent tracking failures, especially during periods of rapid fluctuations of $\util(\ActualGB(t), t)$. Under penalty factor $\lambda(t)$ in Eq.~\ref{eqn:penalty}, these tracking lapses accrue substantial penalties, as the swarm remains anchored to the last known global optimal position with the highest utility instead of following the downward trend in the true optimum, thereby increasing the cumulative error.
\par

SPSO employs subspecies and re-diversification, similar to PSPSO, to track the global optimum in dynamic environments. Unlike PSPSO, SPSO triggers re-diversification only when all subspecies contract within a radius $r_{\mathrm{generate}}$. In our experiments, non-seed particles within subspecies tended to be trapped in local optima, maintaining the spatial spread $\mathfrak{s}_i$ of each subspecies $i$ above $r_{\mathrm{generate}}$. The re-diversification condition was rarely met and the swarm could not refresh its diversity. The tracker subspecies remained stuck in a local optimum, and the computed shift severity $\hat{s}^{(t)}$ progressively decreased. This issue is worsened by SPSO's lack of random perturbations in the subspecies. Because $\hat{s}^{(t)}$ guides the tracker designation to follow the global optimum, the SPSO's tracking capability degrades under these conditions, causing a higher average cumulative tracking error than that of our proposed method.
\par

\begin{table}
\caption{\nngpso tracking errors against particle counts $|p|$}
\centering
\begin{tabular}{lrl}
\toprule
\textbf{Algorithm} & \textbf{$|p|$}&  \textbf{$\mathrm{error}(\bar{\mathcal{E}})$}\\
\midrule
\cnngpso & 5   & $4.860 \pm 3.847$ \\
\cnngpso & 7 & $4.688 \pm 3.749$ \\
\cnngpso & 10  & $4.656 \pm 4.005$ \\
\dnngpso &5 & $4.390 \pm 3.531$ \\
\dnngpso &7 & $4.248 \pm 3.801$ \\
\dnngpso &10 & \textbf{4.129 $\pm$ 3.562} \\
\bottomrule
\end{tabular}
\label{tbl:dnnpso_particle_comparison}
\end{table}

\subsection{Scalability by Particle Count}
To assess scalability with particle count, we evaluated both variants with $|p|\in\{5,7,10\}$ under the same environment settings as in the main experiment (Table~\ref{tbl:env_setting}), except that the environments per group were set to $E_{\it count} = 3$. The environments were fixed across experiments within each group. For each particle count, we report the aggregated mean and SD $\mathrm{error}(\bar{\mathcal{E}})$ over all the environment groups using Eqs.~\ref{eqn:aggregate_mean} and\ref{eqn:aggregate_sd}.
\par

As shown in Table~\ref{tbl:dnnpso_particle_comparison}, both variants exhibited a decreasing total error as the particle count increased relative to the baseline $|p|=5$ used in the main experiment. For \cnngpso, increasing $|p|$ from 5 to 7 reduced the error by 3.53\% and from 5 to 10 by 4.20\%. For \dnngpso, the reductions were 3.22\% at $|p|=7$ and 5.93\% at $|p|=10$. These results indicate that our method scales with particle count, with \dnngpso showing larger gains than \cnngpso, possibly due to independent networks causing unique particle behavior.
\par

Adding more particles to \dnngpso enhances swarm exploration by increasing the diversity and probability of finding optimal positions. However, adding particles to \cnngpso only increases the probability of exploring optimal points without increasing diversity, as particles share the same neural network, and \nngpso does not use random coefficients, which are often used in PSO algorithms to enhance the exploration capability~\cite{canon_pso}.
\par

Owing to the online training in our proposed methods, increasing the particle count increases the potential maximum training count per time step. This issue is amplified in \dnngpso because it includes information from $D_i(t-1)$ in the data propagated to other particles, potentially exponentially increasing the training data per time step for each particle.
\par

\section{Conclusion}
We modeled the distribution and movement of potential survivors in a post-disaster environment using bivariate Gaussian distributions and randomly positioned the significant locations. Owing to the unpredictable nature of post-disaster environments, we assumed the most complex case of a dynamic optimization problem with a constantly changing environment and fewer particles than the number of potential optima. To reliably track the global optimum, we introduced a DNN into the PSO algorithm so that the DNN could learn the environmental characteristics and guide the swarm to the global optimum.
\par

We conducted experiments in environment groups with increasing complexity to test the tracking performance of our methods against baselines. The DNN of our method successfully guides swarms into the global optimum, resulting in the lowest cumulative tracking error compared with the baselines. This suggests that our method can track the global optimum over a given timeframe. This capability can be useful in solving real-world dynamic optimization problems with time critical solutions, such as dynamic task scheduling or portfolio optimization.
\par

We identified several issues to improve our proposed methods. First, although our method exhibits strong adaptability in dynamic environments, its convergence in static environments remains suboptimal. The neural network component can be improved by using alternative observation strategies, such as orbiting or randomly positioned observation points, to provide more diverse environment observation. Furthermore, a sampling strategy can be used for online training to reduce time while improving the results~\cite{Nisar2023GradientBasedMI}. Additionally, different neural network architectures, such as recurrent neural networks (RNNs), may capture temporal correlations that are not fully utilized in the current feed-forward design, potentially improving the performance. Finally, further evaluation is required in more challenging settings, including higher-dimensional problems and additional metrics such as computational time.
\par

\subsection*{Acknowledgments}
This work is partly supported by JSPS KAKENHI grant number 25K03188.

\bibliographystyle{plain}
\bibliography{references}

\end{document}